\documentclass[a4paper,twoside]{article}      
\usepackage{amsmath,amssymb,amsfonts} 
\usepackage{graphics}                 
\usepackage{color}                    
\usepackage{hyperref}                 
\usepackage{comment} 			
\usepackage{enumitem}

\date{\small\it \today}
\title{Discussion on Mechanical Learning and\\Learning Machine
\footnote{Great thanks for whole heart support of my wife Yiping. Thanks for Internet and research contents contributers to Internet.}}

\author{ Chuyu Xiong \\
{\small Independent researcher, New York, USA} \\
{\small Email: chuyux99@gmail.com}
}

\begin{document}

\maketitle
\begin{abstract}
Mechanical learning is a computing system that is based on a set of simple and fixed rules, and can learn from incoming data. A learning machine is a system that realizes mechanical learning. Importantly, we emphasis that it is based on a set of simple and fixed rules, contrasting to often called machine learning that is sophisticated software based on very complicated mathematical theory, and often needs human intervene for software fine tune and manual adjustments. Here, we discuss some basic facts and principles of such system, and try to lay down a framework for further study. We propose 2 directions to approach mechanical learning, just like Church-Turing pair: one is trying to realize a learning machine, another is trying to well describe the mechanical learning.

\end{abstract}

 \\

The only way to rectify our reasonings is to make them as tangible \\
as those of the Mathematicians, so that we can find our error at a glance, \\
and when there are disputes among persons, we can simply say: \\
Let us calculate, without further ado, to see who is right. \\
\indent \hspace{20pt} ----Gottfried Leibniz\\

\section{Introduction}
In recent years, machine learning becomes hot topic of  research and IT development. Yet, there are still some very fundamental problems need to be addressed. In the effort to understand these problems, we brought up the term mechanical learning. Here, we will try to lay down the discussion framework for mechanical learning.
 
While electronic devices can do numerical computation effectively, and actually can do many complicated even intelligent things, however, inside the device there is a core that is very {\it mechanical}, i.e. the device is governed by {\it a set of simple and fixed rules}. The ability of electronic device doing complicated information processing comes from that it is running a pre-installed program that is from human intelligence. In another words, the core of  computing device is very mechanical, and its ability of complicated information processing is endowed by human intelligence.

In machine learning software, the situation seems different. For a machine learning software, its ability of information processing is, at least partially, from learning.  Naturally, we would ask: can a computing device endows itself the ability of information processing by learning? It is not easy to answer. Machine learning software seems acquires ability of information processing from learning, however, if we look more deeply, we would notice that such learning very heavily depends on human intervenes and involvements. This motivates us to consider to isolate the part of learning that does not need human intervene. So, it follows to put the requirement of  {\it a set of simple and fixed rules}.     

The purpose to use term {\it mechanical} is to emphasis {\it a set of simple and fixed rules}, not to mean gears, levers, pushers and pullers. We would like to call doing things by a set of simple and fixed rules as mechanical. This is in-line with historical usage, such as mechanical reasoning and mechanical computing. 

At first, we introduce IPU (information processing unit), and base our discussions on IPU. We then explain more why we are interested in {\it mechanical learning}, or learning by {\it a set of simple and fixed rules}. To demonstrate the effectiveness of such requirement, we show that we could reach some important implication by this line of thinking. 

Once we are thinking in this way, we naturally draw analogy between mechanical computing and mechanical learning. Also, naturally, we recall the fundamental works of Turing and Church on mechanical computing and reasoning. This strongly suggests we should go 2 different and equivalent approaches to mechanical learning: to extend Church-Turing thesis, i.e., one way is to construct an universal learning machine, and equivalently, another way is to well describe mechanical learning. Church-Turing thesis gave people the key to understand mechanical computing, and we believe, the extended Church-Turing thesis will give us good guidance on mechanical learning. 

This paper is the first one for our discussions on mechanical learning. We will write down our studies in next papers. In last section, we put down some topics that we would like to discuss further.

In this discussion, we will restrict us to spatial learning, not consider temporal learning. First, let's make the distinguish of {\it spatial} and {\it temporal} roughly here. For learning machine, we of course concern patterns. Roughly say, the pattern along the incoming space is spatial pattern, while the pattern along the time line is temporal pattern. To say that we restrict us to spatial learning, it means that we only consider the pattern of incoming space, not consider the pattern of several patterns sequentially coming. For a simple example,   consider letters: A, B, C, ..., Z, each single letter is one spatial pattern. To restrict to spatial pattern, the learning machine will be able to consider (and possibly to  learn) patterns A, B, ..., Z, but will not, and is not able to consider (and possibly to  learn) the sequence of letters, such AB, CZW, etc. The meaning of term {\it spatial} will become more clear in later discussions. By restricting us to spatial learning, we can simplify the discussion so that we can go in more depth. Of course, any true learning system should consider both spatial and temporal learning together. But, that is beyond the scope of current discussion.

\section{Information Processing Unit and Mechanical Learning}\label{table}
Here,  we try to formalize the mechanical learning and related principles. We will start from information processing since learning is inseparably linked to information and how information is processed. Actually, information processing is computing. Thus, what we are talking about here is actually to view computing in a different angle. 

{\bf Definition} {\it Information Processing Unit:} \\
{\it Information Processing Unit} (IPU) is such an entity: it has input space and output space, input space is one $N$-bit binary array $I$; output space is one $M$-bit binary array $O$. For any input $i \in I$, there will be one corresponding output $o \in O$. We will call it as $N$-$M$ IPU. 

\begin{center}
\begin{picture}(300,150)(0,0)
\put(0,-208){\resizebox{10 cm}{!}{\includegraphics{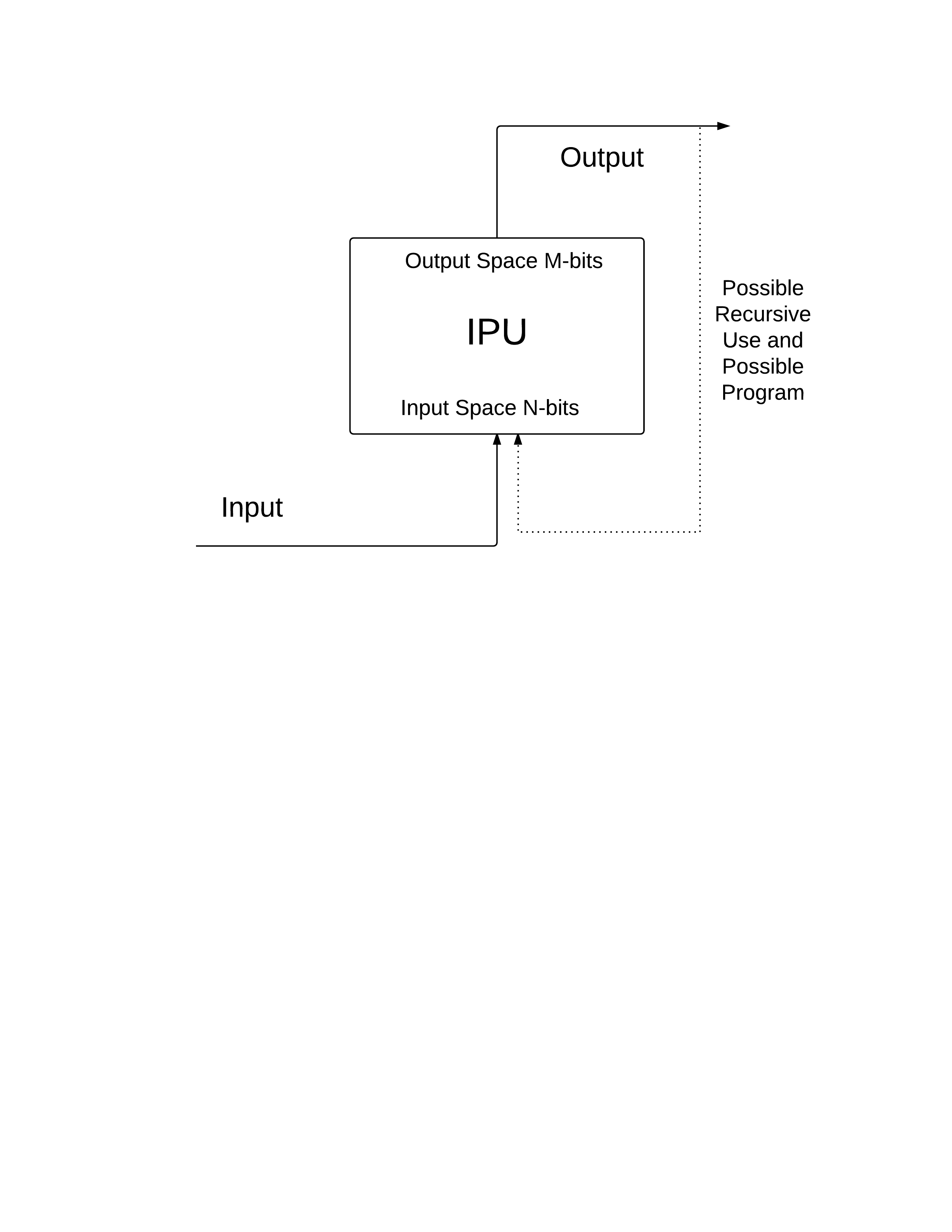}}}
\end{picture}

{\bf Fig. 1. Illustration of $N$-$M$ Information Processing Unit} 
\end{center}

At this moment, we do not focus on how the information is processed inside IPU, instead, we focus on information input and output. In our notation, for one input $i \in I$, the output is $o = P(i)$, we call $P$ the processing of IPU. So, in fact, the processing $P$ is one mapping between binary array, $P: I \to M$, i.e. for every member $i \in I$, there is one $o \in O$, so that $o = P(i)$. Clearly, one particular processing defines the particular behavior of IPU. 

Here, we can make the term {\it spatial} and {\it spatial learning} more clear. Exactly, the meaning of spatial is: for input $i \in I$, the output $o = P(i)$ will only depend on $i$, not on any other, no any context to depend, such as the previous input or later input. So, the term {\it spatial} exactly means only to consider input itself, and no any influence from time line. If the output also depends on context, the IPU would be called temporal. Clearly, we could make one IPU be temporal. But, we will restrict us on spatial IPU. This is the exact mean for {\it spatial}. 

We are not just interested in one particular processing. We are mostly interested in how processing is changing and how processing is learned. Thus, we would consider all possible processing for $N$-$M$ IPU. We have this:

{\bf Lemma 1}  \\
For $N$-$M$ IPU, the total possible processing are $2^{M 2^{N}}$. Using the term of bits, the number of all possible processing is $M 2^{N}$ bits.

The proof of Lemma 1 is very straightforward. But, this simple fact has great implications as we will demonstrate later. Simply say, except very small $N$,  for $N$-$M$ IPU, the number of total possible processing is extremely huge. To show this as one quick example, we can consider the cognition of hand written digits. In this case, we need to consider input space of $28\times28$ black-white pixels. So, the input space has $N = 28\times28 = 784$. Thus, all possible processing are in the order of $2^{784}$ bits!  

IPU actually can be thought as a computing unit. In fact, a computer CPU is one IPU. So, with recursive usage and a well designed program to do the recursive usage, any computation can be done by one IPU. In this sense, it seems no reason to introduce the term of IPU. However, the reason to introduce such term is to emphasize the information processing: input $N$ bits information and output $M$ bits information. In this way, we can focus on properties of information processing, specially how the processing is learned.   

To change the processing of one IPU has many ways. To change the processing inside it manually, either by programming, or some fine tuning, clearly is one way. However, we only are interested in the changes driven by experience/data. If for a IPU $\mathcal{M}$, and if there is a sequence of data $\{ i_k, o_k\}, i_k \in I, o_k \in O, k = 1, 2, \ldots$ feed into $\mathcal{M}$, and under such data driving, the processing of $\mathcal{M}$ is changing, e.g. from $P_1$ to $P_2$, we would say $\mathcal{M}$ is learning. So, we have:

{\bf Definition} {\it Learning of IPU:} \\
For one IPU $\mathcal{M}$, if the processing of $\mathcal{M}$ is changing under the driving force of feed-in of input data and feedback of output data, we call $\mathcal{M}$ is learning. 

For learning of IPU, our focus is: under what data, what changes of processing occur, and how. Yet, there could still have too many things involved in changing of processing. For example, to manually modify software sure could change the processing. More subtly, to manually put bias into the computing system could also change the processing. Surely, we would like to exclude all such factors. So, we have:  

{\bf Definition} {\it Mechanical Learning of IPU:} \\
For one IPU $\mathcal{M}$, if the processing of $\mathcal{M}$ is changing under the driving force of feed-in of input data and feedback of output data, and the changing is according to {\it a set of simple and fixed rules},  we call $\mathcal{M}$ is doing mechanical learning.
 
This definition could hardly be called as a mathematical definition. But, it is the best so far we could make. We will discuss more in later section.

If we can build one computing system that can realize mechanical learning, we call it learning machine. A learning machine could be specialized hardware, or a pure software sitting in a computing environment (cloud, supercomputer, PC, or even cell phone), or combination of them, etc. The most important property of a learning machine is: it is doing mechanical learning. 

One immediate consideration for a learning machine would be: universal learning.  

{\bf Definition} {\it Universal Learning:} \\
For one learning machine $\mathcal{M}$, if for any giving processing $P$ (i.e. one mapping from $I$ to $O$), no matter what the current processing of $\mathcal{M}$ is, $\mathcal{M}$ could learn $P$ (i.e. its processing becomes $P$), then, we call $\mathcal{M}$ universal.

Simply say, a universal learning machine can learning any processing (starting from any processing). This is a very high requirement, however, as we will see later, this seems very high requirement is actually quite necessary.

There is a group of learning machine we should specially notice: {\it parameterized learning machine}.

{\bf Definition} {\it Parameterized Mechanical Learning:} \\
If a learning machine $\mathcal{M}$ is well defined and be controlled by a group of parameters, $\nu_1, \nu_2, \ldots, \nu_L$, and the learning is by changing the parameters, we call such a learning as parameterized mechanical learning.

\begin{center}
\begin{picture}(300,150)(0,0)
\put(0,0){\resizebox{10 cm}{!}{\includegraphics{fig2.png}}}
\end{picture}

{\bf Fig. 2. Illustration of Parameterized Mechanical Learning} 
\end{center}

Currently, almost all machine learning models are actually parameterized. This fact also has big implications as we can see later.

For parameterized learning machine, the learning is actually realized by changing its parameters $\nu_1, \nu_2, \ldots, \nu_L$. Naturally, a question follows: how many possible different processing could be allowed by changing parameters $\nu_1, \nu_2, \ldots, \nu_L$? So, we define: 

{\bf Definition} {\it Effects of Parameter on Processing:} \\
For a parameterized learning machine $\mathcal{M}$, if $\nu$ is one of its parameter, and when $\nu$ varies in its full range, the total possible different processing are less than a number $e$, we then say, this parameter has at most $e$ effects on processing. We often use bits, i.e. $log_2(e)$. 

The relationship between parameters and the effects on processing is very complicated. However, if we know all these parameters have finite values, we at least know the upper limit of total possible different processing. This is true for most computing system. For example, if parameters $\nu_1, \nu_2, \ldots, \nu_L$ are double precision floating numbers, then each $nu_k$ has at most 64 bits finite values, i.e. at most 64 bits effects on processing. This simple fact is also useful.

We talk some examples of IPU in next section. 2-1 IPU is the simplest IPU, yet it still reveals some very interesting properties for us. See appendix for details.

\section{Examples of Mechanical Learning}\label{table}
Now, we see some examples. 

{\bf Examples of IPU} 
\begin{enumerate}
\item See simplest IPU 2-1 IPU in appendix.
\item One mathematical function is one IPU: $P: I \to O$. Such function with $N$ bits variables and $M$ bits function value is one $N$-$M$ IPU. 
\item One software with well-defined input and no context dependence is one IPU. Such software with $N$ bits input and $M$ bits output is one $N$-$M$ IPU. Many statistics software would fit in this category. 
\item One CPU with certain restriction so that it does not have any context is one IPU. Such CPU actually could be viewed as one of mathematical function (but, its definition is complicated). For example, one 64-bits CPU, if we take some restriction, is one 64-64 IPU. 
\item One machine learning software is one IPU. Of course, its processing will be able to change (learn). 
\item Abstractly, and with certain restriction, some processing region in our brain neocortex (for example, that is responsible for digits recognition) can be thought as one IPU ($N$ must be great, and $M$ is small). 
\item Even more abstractly,  and with certain restriction, one decision process is one IPU. Here, the decision process can be in one animal's brain (or even more primary, such as ganglion of a fly), or a meeting of a company's board, etc. For such IPU, $N$ is big, but $M=1$. 
\end{enumerate}

As we see in examples, IPUs are everywhere. Actually, the center of IPU is its ability of information processing. We are most interested in where such ability comes and how such ability adapt/change/learn. Let's see some examples about ability of information processing.

{\bf Examples of IPU, about its information processing}
\begin{enumerate}
\item For IPU formed by a mathematical function, its ability of processing comes from the definition of mathematical function. If this function is computable, we can use computer to realize the processing. So, the ability is from programing.
\item For the software with well-defined input, clearly, the ability is from programming.
\item For CPU, the ability clearly comes from hardware setup and software build into it.
\item For many machine learning software, one would suppose its ability of information processing comes from learning. However, we should examine more deeply, we know that the ability actually partially comes from programming and setup, and effects of both learning and setup are mixed, and not easy to distinguish. This fact actually motivate us to bring up the term of mechanical learning.  
\item For the region in our brain that is responsible for digits recognition, it is safe to claim that the information processing ability is from learning (but a very long learning, starting from baby time, and from school days). We indeed learn this ability. However, the learning is also depends on pre-wiring of the brain region. And we know that the learning is not mechanical. 
\item For one decision process that we abstractly think as one IPU, the ability of information processing is partially from programming, and partially from learning. For example, consider the decision process of a company board as one IPU, then, its ability of information processing partially comes from set up, e.g. the predefined rules, and partially comes from learning, e.g. the success or failure experienced. The learning clearly is not mechanical. 
\end{enumerate}

Of course, we are mostly interested in those IPU, whose information processing is changing, specially, adapting and learning. We can see some examples below.

{\bf Examples of IPU, information processing is changing/adapting/learning } 
\begin{enumerate}
\item For IPU formed by a mathematical function, if the processing can change, then such property must be built in the definition of mathematical function. Mostly likely, it is parameterized. That is to say, $P: I \to O$ is the mathematical function, which has parameters $\nu_1, \nu_2, \ldots$, so that when $\nu_k$ change values, the processing will change accordingly. Learning is to change the parameter values. Actually, many, if not most, IPUs are this type.
\item For so called neuromorphic chip, such as Truenorth of IBM, it can change its information processing. The ability to change the processing is built into the hardware. Such kind of hardware are just at the very beginning of its development, a lot of modification of such chip will be expected. However, we might be able to classify them as parameterized.
\item For one machine learning software, it indeed has ability to change its processing. For most current machine learning software, we can classify them as parameterized.  
\item For a statistics model, it often likes this: mathematical functions + database. This is IPU and its processing is changing/adapting. Database is used to store the incoming data, and mathematical function is statistical model that does calculations based on the data in database. Such IPU is parameterized.   
\item One particular ANN, Restricted Boltzmann Machine (RBM), is the center of many machine learning software. Clearly, it is one $N$-$M$ IPU. RBM is actually completely determined by its entries matrix, a $M$x$N$ matrix. So, it is parameterized.
\end{enumerate}

\section{Why Mechanical Learning?}\label{table}
We defined mechanical learning and learning machine, and saw some examples in the previous sections. Simply say, mechanical learning is: 
One computing system $\mathcal{M}$ improves its ability of information processing according to a set of simple and fixed rules under the driving of incoming data. 

But, why are we interested in  {\it a set of simple and fixed rules}? Let's first explain our thoughts about this.

Seems many current machine learning software are doing well, they do not emphasis mechanical side, but, they emphasis how to make computing system learning from data, and how to do so better. This is perfectly fine. So, is it necessary to bring up the term mechanical and post mechanical requirement on learning?

Against such thought, Jeff Hawkins gave a very strong point \cite{jhaw3}: In order to build a learning machine that has potential to become next generation computing system, it must be independent from the learning tasks. Recall history of computing could help us to better see this. Before von Neumann architecture of computer, there were many systems or devices that could do effective job for certain tasks. However, all of them disappeared. Requirement of "independent from any particular task" is indeed playing the crucial role. Armed by this history knowledge, we would expect to see similar for learning.  

However, current machine learning software heavily depend on human intervenes and involvements, and are quite depend on specific learning tasks. This motivates us to consider to isolate the part of learning that does not need human intervene, and independent from learning tasks. For this reason, we post the mechanical requirement.

Such a thought is not new. Many people have been trying to do so. Numenta developed CLA algorithm trying  to closely simulate human brain neocortex \cite{jhaw2}. By doing so, it hopes to establish one computing system that is independent from any particular learning tasks. Though, at current stage, Numenta's CLA focuses on temporal learning. We think that spatial learning should be studied first and it is easier to deal with spatial learning first. Nonetheless, CLA is an algorithm formed by a set of simple and fixed rules. Once it is setup, human intervene is not necessary and CLA is learning from incoming data. In this sense, we can say, CLA is doing mechanical learning. Of course, CLA is still at its first stage of development, and might not fully realize its goal. However, at least, this is intention of CLA. 

Besides CLA, there are other efforts trying to build master algorithm independent from particular task. For example, Pedro Domingos is trying to unite 5 kinds of learning methods \cite{pedro}: logic learning, connectionist learning, probabilistic learning, analogy, and evolution. If anyone can successfully unites these learning methods, the underneath principle of new method must be simpler, not more complicated. So, we should expect a simple and fixed rules underneath those different types of methods. 

Even more, people now start to question if we can capture the mathematical theory of human brain (of course including learning). For example, see the famous 23 problems of DARPA \cite{darpa}. 

Naturally, in order to do those tasks list above that is aiming very high, we can expect to consider first step: mechanical learning. If we could understand mechanical learning better, we are better prepared for those high tasks.

Now, we can come back to the definition of mechanical learning we gave in section 1. We have to say, it is not very precise. What is mean for "a set of simple and fixed rules"? But, perhaps, this is the best we can do up to now, we cannot give a better and more precise definition for mechanical learning. However, on the other side, it is very important for us to post mechanical requirement on learning, even though we do not know exactly this requirement really means. We can sense the importance of such requirement and can only roughly grasp some basic skeleton of such requirement. 

Again, in order to help us to see better, we will consult history of mechanical reasoning and mechanical computing. It is Leibniz first requested "mechanical reasoning". After him, great amount of efforts were paid to concretely realize "mechanical reasoning", from Cantor, Fred, Hilbert, Russell, till Church and Turing. After many great works done by great scientists and mathematicians, now, we know exactly what mechanical reasoning and computing means: It is what Turing machine does, or equivalently, it is what $\lambda$-calculus describes. It is this great process of pursuing to understand mechanical reasoning and mechanical computing gives us the key to modern computer era. 

We see strong analogy between mechanical computing and mechanical learning. So, for mechanical learning, we can fully expect similar: we do not know exactly mathematical definition of mechanical learning, but, it will be productive if we post the mechanical requirement on learning. By pursuing such requirements, we can propel us to the fully understanding of mechanical learning. This pursue could be a long journey and it might not be easy. But, we can expect the time span is much shorter since we already have the guidance of history of development of mechanical computing and mechanical reasoning.

Currently, a lot of efforts are put on how to do machine learning, and how to do machine learning better. But, in the process, some very fundamental questions have to be addressed. We can list some here:
\begin{enumerate}
\item  What is really learned? What is really learned in a deep learning software?  This question can not be answered precisely. What is really learned in a probabilistic learning module? Is it just some parameter adapting? The question can not be answered precisely. 
\item  What could be learned by computing system? And what could not be learned by computing system? No precise answers.
\item  Why connectionist view is fundamentally important?
\item  Can we integrate logic learning, connectionist learning, probabilistic learning and analogy together effective and smoothly? And how?
\item  How to establish one architecture of learning machine independent from individual learning task, so that this architecture will guide us for next generation of computing?
\item  How to teach a computing system for certain tasks, instead of programming it? Or can we do so? If we can, what is the efficient teaching language/method?
\end{enumerate}

We think, by putting mechanical requirements on learning, we are actually starting to address these fundamental questions, at least from certain point of view, view of rigorous mathematical reasoning. To demonstrate this, we will go following arguments, which is quite simple, but reveal some important implications.

From Lemma 1, we know $N$-$M$ learning machine, the number of all possible processing is $M 2^{N}$ bits. We can have a lemma for parameterized mechanical learning.

{\bf Lemma 2}  \\
For a parameterized learning machine $\mathcal{M}$, if its parameters are $\nu_1, \nu_2, \ldots, \nu_L$,  and each parameter has at most $e$ bits effects on processing of $\mathcal{M}$, then $\mathcal{M}$ could at most have $e \times L$ bits many different processing. In another words, $\mathcal{M}$ at most could learn $e \times L$ bits processing.

The proof is very simple. By combining Lemma 1 and Lemma 2 together, we then have:

{\bf Theorem 1}  \\
For a parameterized learning machine $\mathcal{M}$, most likely, it is not universal learning.

The proof is short: Number of total possible processing of $N$-$M$ learning machine is in order of $M2^N$ bits. Most likely, lemma 2 could apply to $\mathcal{M}$, so the number of processing that $\mathcal{M}$ could learn at most in order of $e \times L$ bits. Thus, unless $L$ is in the order $2^N$, $e \times L \ll M2^N$. It means what  $\mathcal{M}$ could learn is much less than $M2^N$, so $\mathcal{M}$ could not be universal. But, it is extremely unlikely, one parameterized learning machine $\mathcal{M}$ could have such a large group of parameters (even it has, how it can learn?).

Actually, in simple words, Theorem 1 tells us, in order to build an universal learning machine, we could not use parameterized learning machine. This simple fact indicates that almost all current machine learning models are not candidate for universal learning machine. Unlike most of them, CLA of Numenta \cite{jhaw2} might be a system that is not parameterized. However, no one has made a proof yet.   

The arguments above are very simple and shallow. However, it already gives us some strong and very useful indications. Thus, we have strong reason to believe, along this path, efforts could be very fruitful.   

\section{How to Approach Mechancial Learning}\label{table}
How to approach and study mechanical learning? This is not an easy question. However, fortunately, we have a better guide than pioneers of computing. We can recall history of computing to gain some invaluable guidance. 

Before modern computing, people thought about mechanical reasoning and mechanical computing for many hundred years. In fact, people made many devices for such purposes, from ancient abacus, to tablet machine, even to Babbage's mechanical computer. And, on theoretical side, people are fascinated about the mechanical aspects human thoughts, especially computing, and wonder how to explore and use such aspects. Such  thoughts motivated many great scientists and mathematicians working in this direction. And, big block knowledges are accumulated, such as mathematical logic. 
 
But, until Turing and Church, thoughts were very scattered and not systematic, computing devices were designed for special purpose, and without guidance of well developed theory. Simply, people still did not know well what is mechanical logic reasoning and mechanical computing. It is Turing and Church's theory laid down the foundation and let people start to fully understand what mechanical computing is, and how to design universal computer that can do all possible mechanical computing. 

We might express Church-Turing theory in this way: While Turing machine gives one real model of mechanical computing, Church's $\lambda$-calculus gives a very precise description on objects that is mechanically computable. Church-Turing thesis tells us: all mechanically computable (i.e. computable by Turing machine) can be well described by $\lambda$-calculus, and vise versa. 

Using such Church-Turing pair as a guide, in parallel, we will propose to go the same line of thoughts: we should work on 2 equivalent directions: one is trying to establish a real learning machine that is based on a set of simple and fixed rules, and this learning machine can do universal learning; another is trying to well describe the objects that can be learned mechanically, and exactly how the mechanical learning is doing. Going to 2 equivalent directions, would be more fruitful than just going one. For example, if through the well description, we understand that a learning machine should behave in certain way, then, such information will help us to design a learning machine. 

The second direction, could help us to establish teaching language/method to teach a learning machine. We would vision, just like programming is super important for computer, teaching would be super important for learning machine. In another words, instead of programming a machine, we will teach a machine. But, effective teaching needs good teaching language/method besides data. A well description of mechanical learning could guide us to develop such teaching language, just like $\lambda$-calculus guided us to develop programming language. Data is important for teaching. But, teaching language/method are equally important, if not more. 

In this way, Church-Turing pair will continue and be extended: we have a universal learning machine, and we teach the universal learning machine with the well developed teaching language/methods and data.

This is what we propose to do. Actually, we did some works in both directions. We will write down them in different places.

\section{About Further Works}\label{table}
Current article is the first one for our discussions on mechanical learning and learning machine. We will continue to work on the 2 directions talked in last section. We would like to list some topics here. We would be very glad to see more studies on these topics from all possible point of view. 

{\bf About Building Universal Learning Machine} \\ 
In order to build one universal leaning machine, we think following topics are important and fruitful.

{\bf 1.} {\it What way could achieve universal learning?}  As we discussed, any parameterized learning could not be universal. However, how can we avoid parameterized? This is not easy at all. If we use well known mathematical function or sophisticated software as the foundation of the learning machine, it would inevitable become parameterized, since all such mathematical functions and software are all parameterized. We think, from this point of view, connectionist view becomes important. 

{\bf 2.} {\it  First spatial learning, then transit to temporal learning.} Here, for the purpose of simplification, we only discuss spatial learning. However, temporal learning is absolutely necessary. We should first study fully spatial learning, then, armed with the knowledge and tools from such studies, we move to temporal learning, and spatial and temporal learning together. We guess, the transit might not be super hard. After all, we can gain inspiration from human brain. Human brain surely can handle spatial and temporal in uniformly way. This indicates to us, in mechanical learning, there could be a way to handle spatial and temporal learning in  uniformly way. True understanding of spatial learning could be the very key to temporal learning, and vise versa. We have high hope on this part.

{\bf About Descriptions of Mechanical Learning} \\
To well describe mechanical learning, we can list some areas below.  

{\bf 1.} {\it  Generalization.}  Generalization is very desired for a learning machine. That is to say, only need to teach some things to the learning machine, then the learning machine could automatically and correctly generalize to more things. Can a mechanical learning machine to do so? Why this seems intelligent behavior can be achieved by a mechanical learning machine? And how? We sure would like to go in depth for this question. Actually, this is the exactly reason that we propose to study spatial learning first. 

{\bf 2.} {\it  Abstraction.}  As generalization, abstraction is also very desired for a learning machine. Many researchers have already thought that abstraction might be the key of further development of machine learning. At the first step, we need to find some way to well describe abstraction in mechanical learning.  

{\bf 3.} {\it  Prior knowledge and continue learning.}  A learning machine could have prior knowledge. Yet, what exactly is prior knowledge? In what form prior knowledge is in a learning machine? Can and how we inject prior knowledge to a learning machine? How prior knowledge play in the learning/teaching process?   

{\bf 4.} {\it Pattern Complexity vs. Capacity of Learning.}  Very naturally, learning is closely related to patterns. We can intuitively say, more complex the pattern associated to learning is, harder the learning would be. But, is it so? If it is so, can we measure the complexity and hardness exactly? Also, intuitively, we can think that if one learning machine has a better capacity of learning, it can learn more complex things. But, is it so? If so, can we say more exactly?

{\bf 5.} {\it Teaching, training and data.} For learning machine, programming could still be a way to make it to do desired tasks. But, teaching or training would be more important and more often be used. So, how to do teaching or training efficiently and effectively? Should we have to use big data? If so, what big data is really used for? 

{\bf About Integration of Different Types of Machine Learning} \\
Pedro Domingos listed 5 kinds of learning methods \cite{pedro}: logic, connectionist, probabilistic, analogy, and evolution. All of them have sounding supports, and each is doing better than others in some areas. This indicates each of them indeed stands on some important part of the big subject: learning. Naturally, it is best to integrate them, instead of to choose some and discard others.   

We would think connectionist view is going to play a central role, since it is very hard to imagine logic view could integrate connectionist view (specially, not parameterized), but conversely, it would be easier to imagine (though, we do not know how at this time).  Also, it might be easier to imagine that a connectionist model can handle analogy. Can we imagine such a system: it is a connectionist system, and inside it, it accomplishes logic view, probabilistic view and analogy naturally, and evolution is helping this system improving? If we can achieve such a system, or at least partially, we would progress very well.

\section*{Appendix}\label{table}
{\bf 2-1 IPU } 

2-1 IPU is the simplest IPU. For a 2-1 IPU, there are totally 16 ($2^{2^2} = 16$) possible processing. We can see all processing in following value table.

\begin{center}
\begin{table}[h]
\centering
    \begin{tabular}{|c|c|c|c|c|c|c|c|c|c|c|c|c|c|c|c|c|}
        \hline
        ~     & $P_0$ & $P_1$ & $P_2$ & $P_3$ & $P_4$ & $P_5$ & $P_6$ & $P_7$ & $P_8$ & $P_9$ & $P_{10}$ & $P_{11}$ & $P_{12}$ & $P_{13}$ & $P_{14}$ & $P_{15}$ \\ \hline
        (0,0) & 0     & 1     & 0     & 0     & 0     & 1     & 1     & 0     & 1     & 0     & 1        & 1        & 1        & 0        & 0        & 1        \\ 
        (1,0) & 0     & 0     & 1     & 0     & 0     & 1     & 0     & 1     & 1     & 1     & 0        & 1        & 1        & 0        & 1        & 0        \\ 
        (0,1) & 0     & 0     & 0     & 1     & 0     & 0     & 1     & 1     & 1     & 1     & 1        & 0        & 1        & 1        & 0        & 0        \\ 
        (1,1) & 0     & 0     & 0     & 0     & 1     & 0     & 0     & 0     & 1     & 1     & 1        & 1        & 0        & 1        & 1        & 1        \\
        \hline
    \end{tabular}
\end{table}

{\bf Tab. 1.  Value table of all processing of $2$-$1$ IPU} 
\end{center}

Some processing are quite familiar. For example, $P_7$ is atually XOR logical gate, $P_9$ is OR logical gate, $P_4$ is AND logical gate. Also note, $P_8$ is flip of $P_0$, $P_9$ is flip of $P_1$, etc.

$P_2$, $P_3$ and $P_4$ are most important, which are the building blocks in 2-1 IPU.  $P_0$ is for processing that output is always 0 no matter what input is. $P_1$ looks not for real. But it is for completeness of discussion. It is easy to see that the rest of processing can be constructed by the above. For example, $P_{10} = P_1 + P_3 + P_4$. 

Above, we tell what 2-1 IPU is. And, we point out that we can design an effective learning methods so that all processing could be learned. For 2-1 IPU, this is very simple. However, this simplest case could still give us some good guide. For example, 2-1 IPU is embedded in any IPU. Therefore, any learning machine should effectively handle all processing we listed above, at least.


\begin{thebibliography}{99}

\bibitem{jhaw2} Jeff Hawkins. White Paper: Cortical Learning Algorithm, Numenta Inc, 2010. \htmladdnormallink{http://www.numenta.org}{http://www.numenta.org}

\bibitem{jhaw3} Jeff Hawkins. Talk on Numenta Software.  \htmladdnormallink{http://www.numenta.org}{http://www.numenta.org}

\bibitem{darpa} The world's 23 toughest math questions: DARPA's math challenges. The question 1. \htmladdnormallink{http://www.networkworld.com/community/blog/worlds-23-toughest-math-questions}{http://www.networkworld.com/community/blog/worlds-23-toughest-math-questions}

\bibitem{pedro}  Pedro Domingos. The Master Algorithm, Talks at Google.  \htmladdnormallink{https://plus.google.com/117039636053462680924/posts/RxnFUqbbFRc}{https://plus.google.com/117039636053462680924/posts/RxnFUqbbFRc}

\end{thebibliography}
\end{document}